___________________________________________________________________________________________________________________

# An Effective Pixel-Wise Approach for Skin Colour Segmentation
## Using Pixel Neighbourhood Technique

Tejas Dastane, Varun Rao, Kartik Shenoy, Devendra Vyavaharkar

Department of Computer Engineering  
K.J. Somaiya College of Engineering, University of Mumbai  
Mumbai, India

*Abstract*—This paper presents a novel technique for skin colour segmentation that overcomes the limitations faced by existing techniques such as Colour Range Thresholding. Skin colour segmentation is affected by the varied skin colours and surrounding lighting conditions, leading to poorskin segmentation for many techniques. We propose a new two stage Pixel Neighbourhood technique that classifies any pixel as skin or non-skin based on its neighbourhood pixels. The first step calculates the probability of each pixel being skin by passing HSV values of the pixel to a Deep Neural Network model. In the next step, it calculates the likeliness of pixel being skin using these probabilities of neighbouring pixels. This technique performs skin colour segmentation better than the existing techniques.

*Keywords*-*Skin Detection; HSV Colour Space; Deep Learning; Naïve Bayes; Decision Trees; Pixel Neighbourhood; Pixel-Based Segmentation; Region-Based Segmentation; Skin Colour Segmentation*

__________________________________________________*****__________________________________________________

## I. INTRODUCTION

Skin colour segmentation is the process of identifying the skin-coloured areas in an image or video [1]. It is an essential pre-processing step for applications such as Hand Detection, Facial Recognition, Human Computer Interaction, Facial Expression Analysis, Security Systems and Gesture Tracking. It is also crucial in eliminating the insignificant background details and keeping only the relevant information for further processing. There are two types of skin colour segmentation techniques:

- Pixel-based segmentation - Each pixel is evaluated based on certain conditions to determine whether it is a skin or non-skin pixel [2].
- Region-based segmentation – The spatial relationship of pixels is considered to identify skin regions. An initial skin region is expanded by adding more pixels that satisfy the criteria for skin colour [2].

Usually, the method of Colour Range Thresholding is used for skin colour segmentation. Other techniques, involving the use of Neural Networks and Machine Learning algorithms, are also used [3]. They are time-intensive and thus, unsuitable for real-time applications [4], but optimisations can make them faster.

The purpose of this paper is to introduce an effective technique to perform skin colour segmentation. The 2<sup>nd</sup> section of the paper introduces widely used techniques such as Colour Range Thresholding. It also explains their drawbacks. In the 3<sup>rd</sup> section, we introduce our new technique. The 4<sup>th</sup> section presents the experimental results and discusses the findings after application of our proposed technique. Further research that could be carried out is explained in the 5<sup>th</sup> section and 6<sup>th</sup> section provides a conclusion derived from interpreting the results.

## II. RELATED WORK

Various techniques have already been developed to achieve Skin Colour Segmentation. The most common and simplest one is segmentation based on thresholds in HSV or RGB or YCbCr colour space or any combination among these. Skin Colour Range Thresholding involves first selecting the colour space(s) and then determining the criteria of thresholding [1]. For every colour space, there exists an optimum skin colour range criterion such that the resulting accuracy will approximately be the same in all the models [5].

In YCbCr colour space, the skin colour range will vary depending on the luminance. If we use a definite range for low, medium and high luminance, we will get poor results. This is because in YCbCr space, skin colours' cluster has a highly irregular boundary [3]. Hence, Phung *et al.* [3] have used K-Means clustering and have divided this cluster into three Gaussian clusters. Now, the pixel to be classified is tested to see whether its distance to the closest cluster is less than a pre-decided threshold. If yes, then it is classified as skin, else it is classified as non-skin. This technique is subject to these limitations [2]:

- Skin colour varies from person to person.
- The image on which segmentation is to be done may contain objects with colour closer to the skin colour range.
- Skin colour as seen in image depends on illumination.
- The threshold is difficult to determine and usually determined empirically instead of using theory or pure logic [6]. This threshold is determined usually by histogram analysis [7]. Even if we do get the threshold, the decision boundary in the colour space would be cubic and not curved.
- Provides poor segmentation results in relatively low light conditions.

182

___________________________________________________________________________________________________________________

Segmenting the image into skin and non-skin objects is usually not enough since the background may also have skin-like objects. Hence, this threshold-based methodology is usually combined with motion, texture and edge features [2, 4].

Other techniques employed for skin colour segmentation include use of classification and clustering algorithms such as Bayesian classifiers, Gaussian classifiers, Neural Networks [3]. These algorithms prove to be more effective than Colour Range Thresholding but are also prone to generating noise by classifying a small group of isolated pixels as skin, when they are not.

### III. THE PIXEL-NEIGHBOURHOOD TECHNIQUE

*A. Methodology*

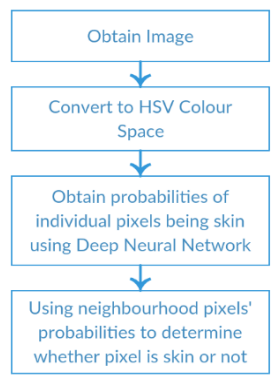

Figure 1.  Flow diagram for the Pixel-Neighbourhood Technique

The technique was inspired by the fact that we encounter a group of skin pixels. It is not possible to find a single skin pixel surrounded by all non-skin pixels. This technique uses the pixel's neighbourhood to classify the pixel as skin. It is a two-step process as represented in Fig. 1. A pixel can be judged on the following conditions:
1) Pixel is of skin colour.
2) Whether there are skin pixels in its neighbourhood.

The pixels that satisfied both these conditions are classified as skin. To judge the pixel on the first condition, a model is used that gives the following two outputs:
1) The probability of the pixel being skin. (Denoted as P(colour=skin))
2) The probability of the pixel being non-skin. (Denoted as P(colour=non-skin))

Since having skin pixels in a pixel's neighbourhood increases the likeliness of the pixel being skin, the likeliness was then calculated to judge the pixel on the second condition. Let it be denoted as L(pixel=skin). Similarly, L(pixel=non-skin) was calculated. The final probabilities are then found as:

P(pixel=skin) = P(colour=skin) * L(pixel=skin)

*B. Colour Space Selection*

We need to represent the dataset in a specific colour space before we apply this technique. This is because different colour spaces have different forms of representing colours. RGB represents colours as a mixture of Red, Green and Blue colours. HSV represents it as a combination of Hue, Saturation and Value. We experimented with two colour spaces for the dataset – RGB, HSV. With the RGB colour space, the data would be sensitive to changes in intensity. For real-time skin colour segmentation, there will be changes in intensity regularly and might affect segmentation. Whereas in HSV colour space, V stands for Value and signifies intensity in the image. Using HSV colour space, a model would be able to learn the intensity range as well as the underlying colour, independent of intensity in the skin colour range. We decided to use HSV colour space.

*C. Dataset used*

We used the UCI skin segmentation dataset [8] for training the classifier. The data contains skin pixels from face samples of people of various skin tones. It also contains pixels which do not lie in the skin colour range. The data was compiled from the FERET database and PAL database. To improve skin classification on Indian skin colours, we appended our own data to this dataset, compiled by extracting skin colours from samples of our hands. Currently, our dataset contains 299,629 samples. It contains 70,960 samples for skin colour values and 228,669 values for pixel colour other than skin colour. The dataset was divided into a training and testing set of 209,740 samples and 89,889 samples respectively.

*D. Model Selection*

Various models were used to judge a sample pixel on the first stated condition. Each model returns an array of two values – Probability of pixel being skin and probability of pixel being non-skin. In section 4, we will briefly summarise the results of these models and determine which technique is most suitable for judging pixel on the first condition.

*1) Deep Neural Network*

Artificial neural networks are computing systems which are inspired by the neural networks in animal brains. These networks learn tasks based on examples, called as the training set, rather than being programmed specifically for each task. In our case, the task is to classify the pixel as a skin pixel or a non-skin pixel.

Neural networks consist of one input layer, one output layer, and one or more hidden layers in between the input and the output layers. These layers consist of units called as neurons. The hidden layer contributes to modelling non-linear relationships in the data.

For our task of skin pixel classification, we are using a deep neural network with an architecture as shown in Fig. 2.

We have implemented the above Neural Network model using Keras library [9]. Our model will be using the categorical cross-entropy loss function, along with the Adam optimizer using the constants specified in the Keras documentation (https://keras.io/optimizers/), and the original paper on Adam [7]. We will be training our model for 12 epochs, with a batch size of 53. Since our model consists of a good number of hidden layers, we shall refer to our neural network as 'Deep Neural Network' in the following discussion.

*2) Naïve Bayesian model*

We created a probability model based on the Bayesian principle to classify pixel colour as skin and non-skin. We experimented with different bin sizes on the data during the building phase of the model. The best results were obtained when the data was not divided into bins. A bin can be defined as a range of values, e.g. 0-5. The model is built directly using

183

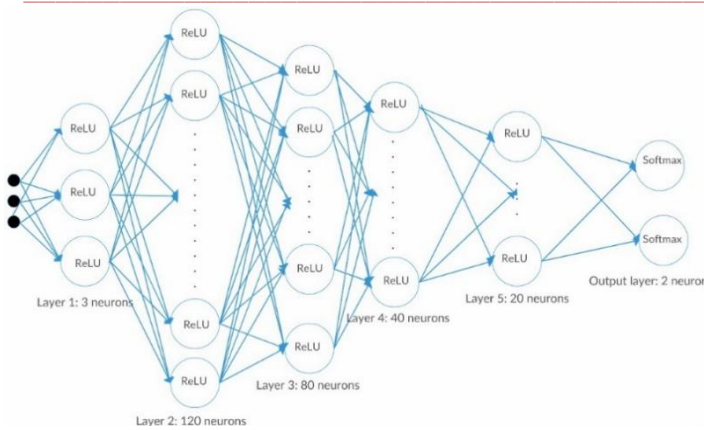

Figure 2.  Deep Neural Network Model Architecture

the normalized H, S & V values without dividing data in bins. For each unique attribute value, the number of tuples corresponding to each class (skin or non-skin) are counted. These counts are then converted into probabilities by dividing the total number of tuples in the training set corresponding to each class value. Pixel is classified as 'skin' or 'non-skin' using the Bayesian principle:

$$P(X \mid (H, S, V)) = \frac{P((H,S,V)|X) * P(X)}{P((H,S,V))}$$

(H, S, V) is the tuple to be tested and X is the class upon which it is tested. This equation is tested for each class and the pixel is classified as the class which gives us highest probability.

*3) Decision tree*

Decision Trees are simple tree-like graphs or models of decisions and their possible outcomes. Although used heavily in operational research, they are a popular tool in machine learning because of their ease in understanding. Decision Trees are like flow-charts. In these flow-charts, each internal node represents the test on an attribute, each branch represents the outcome of this test, while each leaf node represents a class label. For constructing a decision tree, every-time when a node is created, the attribute on which the test should be conducted is decided. This is done using the Gini Index, or the Information Gain or other ways to choose the attribute. For finding results of decision trees on our data set, we have used the Scikit-learn library [10] which contains an inbuilt function for creating decision trees by taking in the raw data.

*E. Calculating likeliness*

As mentioned before, having more skin-like neighbour pixels increases the likeliness of pixel being skin, thus the likeliness is dependent on P(colour=skin) of each neighbourhood pixel. Thus,

$$L(pixel=skin) \alpha \frac{1}{c}\sum_{i=1}^{c} P(color_i = skin) \quad (1)$$

$$L(pixel=non\text{-}skin) \alpha \frac{1}{c}\sum_{i=1}^{c} P(color_i = non - skin) \quad (2)$$

Where $P(colour_i=skin)$ represents probability that a neighbourhood pixel lies in the skin colour range and 'c' represents the number of neighbouring pixels. Denoting –

$$L_1 = L(pixel=skin),$$
$$L_2 = L(pixel=non\text{-}skin),$$
$$S_1 = \sum_{i=1}^{c} P(color_i = skin),$$
$$S_2 = \sum_{i=1}^{c} P(color_i = non - skin)$$

we can write Equations 1 and 2 as –

$$L_1 = \frac{k_1}{c} * S_1 \quad (3)$$

$$L_2 = \frac{k_2}{c} * S_2 \quad (4)$$

Where $k_1$, $k_2$ are constants, c is the number of neighbourhood pixels. By adding Equations 3 and 4, we can write–

$$L_1 + L_2 \alpha \frac{1}{c} * (S_1 + S_2) \quad (5)$$

Since sum of $S_1$ and $S_2$ would be c,

$$L_1 + L_2 = K$$

To maximize detection of skin, we derived an equation for $k_1$ and $k_2$ by combining Equations 3,4 and 5:

$$k_1 = \frac{K*c}{S_1}, k_2 = 0$$

In our implementation, we have considered K=1. Value of $k_1$ is set to maximum (above equation) when we classify a pixel as skin. When a pixel gets classified as non-skin, we set it to 1. Using this technique, we then calculate L(pixel=skin) and find the final probability.

IV.  RESULTS

*A. System and software used for experimentation*

For finding the effectiveness of these techniques, we have tested our hypothesis using a Lenovo Yoga 500. The laptop uses an x64-based Intel Core i7-5500 CPU, clocking at 2.40 GHz and a RAM of 8 GB. The Operating System used was the 64-bit version of Windows 10 Home. It was implemented in the Python language. The Decision Tree was constructed using the Scikit-Learn package of Python [10], while the Deep Neural Network for training the model was built using the Keras API [9] over TensorFlow [11]. To test our models on an actual image, we have used the OpenCV library [12] on Python. For training and testing the models, we have split our dataset into the training set and the testing set, with the testing set being 30% of the entire dataset. All the statistical measures of performance below have been found out on the testing set.

*B. Experimental Results*

TABLE I.   STATISTICAL MEASURES OF PERFORMANCE FOR THE MODELS.

| Parameter | Models | | |
|---|---|---|---|
| | *Deep Neural Network* | *Naïve Bayesian model* | *Decision Trees* |
| Accuracy | 97.32% | 93.23% | 96.35% |
| Sensitivity | 93.68% | 90.17% | 93.14% |
| Specificity | 98.41% | 94.17% | 97.33% |
| Precision | 94.74% | 82.58% | 91.45% |
| F-1 score | 94.2% | 86.20% | 92.29% |

| Parameter | Models | | |
|---|---|---|---|
| | *Deep Neural Network* | *Naïve Bayesian model* | *Decision Trees* |
| AUC | 0.9604 | 0.9217 | 0.9524 |

TABLE II.　　CONFUSION MATIX FOR DEEP NEURAL NETWORK

| N=89889 | *Actual=Skin* | *Actual=Non-skin* | *Predicted* |
|---|---|---|---|
| Predicted=Skin | 19745 | 1097 | 20842 |
| Predicted=Non-skin | 1333 | 67714 | 69047 |
| Total | 21078 | 68811 | 89889 |

TABLE III.　　CONFUSION MATIX FOR NAÏVE BAYES CLASSIFIER

| N=89889 | *Actual=Skin* | *Actual=Non-skin* | *Predicted* |
|---|---|---|---|
| Predicted=Skin | 19005 | 4010 | 23015 |
| Predicted=Non-skin | 2073 | 64801 | 66874 |
| Total | 21078 | 68811 | 89889 |

TABLE IV.　　CONFUSION MATIX FOR DECISION TREE

| N=89889 | *Actual=Skin* | *Actual=Non-skin* | *Predicted* |
|---|---|---|---|
| Predicted=Skin | 19632 | 1836 | 21468 |
| Predicted=Non-skin | 1446 | 66975 | 64821 |
| Total | 21078 | 68811 | 89889 |

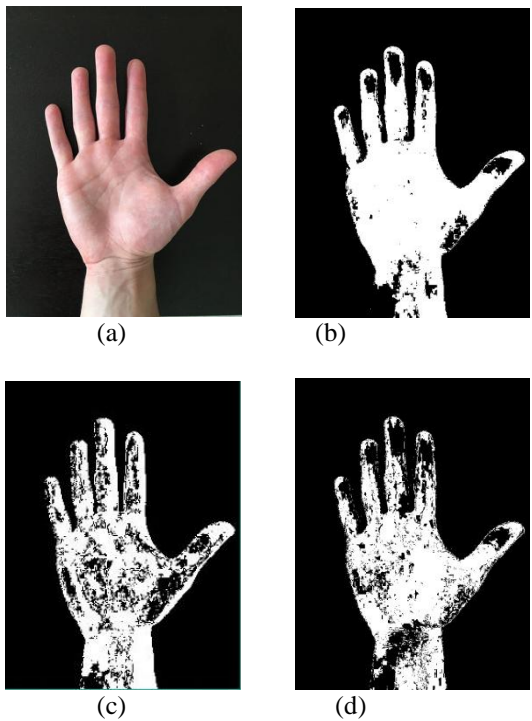

Figure 3. Results after applying various models to a sample image: (a) Original Image (b) Segmentation using Deep Neural Network (c) Segmentation using Naïve Bayes Classifier (d) Segmentation using Decision Tree

We can observe from the Statistical measures in Table 1 that Deep Neural Network model has the highest testing accuracy when classifying skin pixels. The confusion matrices in Tables 2, 3 and 4 further elaborate how many samples were detected correctly as well as incorrectly. We also experimented these models on an image and results are as shown in Fig. 3(b), 3(c) and 3(d). This image is of size 450x600 pixels. The results clearly show that the Deep Neural Network model performs the best skin colour segmentation on a sample image and shows highest accuracy. We thus decided to use the Deep Neural Network model for determining whether a sample pixel lies in skin colour range.

We now represent the results after using Pixel Neighbourhood technique. Fig. 4(d) shows the results of using the Pixel Neighbourhood technique using probabilities obtained by the Deep Neural Network classifier. The sample used here is 450x600 pixels in resolution. Without using any optimisation, the result took approximately 12s for segmentation.

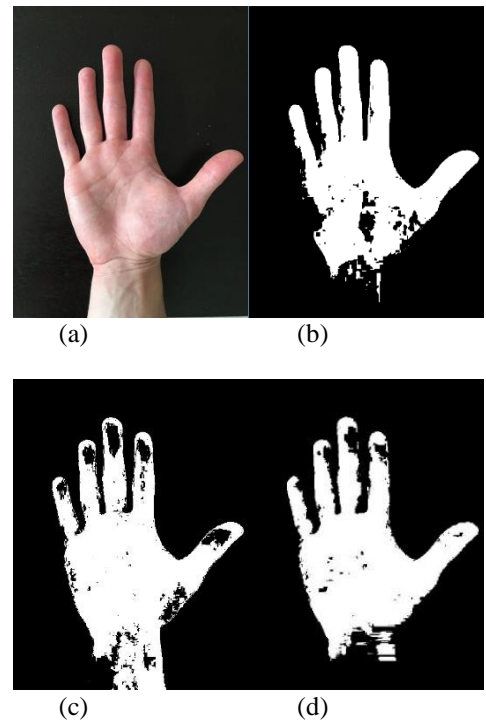

Figure 4. Results after applying various models to a sample image: (a) Original Image (b) Segmentation using Colour Range Thresholding(c) Segmentation using Deep Neural Network (d) Segmentation using Pixel Neighbourhood Technique with Deep Neural Network classifier

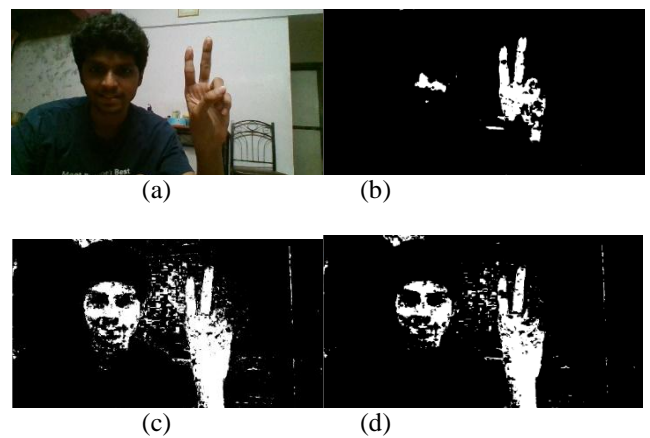

Figure 5. Results after applying various models to a sample image: (a) Original Image (b) Segmentation using Colour Range Thresholding (c) Segmentation using Deep Neural Network (d) Segmentation using Pixel Neighbourhood Technique with Deep Neural Network classifier

185

The sample image in Fig. 5 has a resolution of 1280x720. It can be observed that in Fig. 4(b), Colour Range Thresholding provides better skin colour segmentation. However, in Fig. 5(b), we can observe that the skin colour segmentation is very poor, omitting parts of the person's face. In both cases, YCbCr colour space conversion was used before pixel classification. Pixels lying in the YCbCr range (0,147,60) and (255,180,127) were classified as skin. Thus, it is imminent that Colour Range Thresholding provides good skin colour segmentation under good lighting conditions only and fails when light is low. As observed in Fig. 4(c) and Fig. 5(c), Deep Neural Network classifier was able to classify enough pixels as skin in any condition. We can conclude from the results obtained from using our Pixel Neighbourhood Technique in Fig. 4(d) and Fig. 5(d) that the noisy pixels classified as skin were reduced, skin areas got denser and sparser areas were reduced. Especially in Fig.4(d), it is more clearly seen that it fills some black areas and reduces sparse areas of skin at the bottom of Fig. 4(c).

## V. FUTURE WORK

There are still ways to make this technique perform better. An effort was made to improve the currently available UCI dataset to perform better on Indian skin colours and in relatively low light conditions. The accuracy of this technique can be further improved by including a wider variety of skin and non-skin colour samples.

It can also be noted that the model used in this technique consumes more time than existing techniques. Optimisations such as resizing image to half its size for segmentation and then again resizing the mask generated to original size can boost the performance of this technique by a large factor. Other optimisations can also be worked upon. Optimisations such as performing parallel computations or usage of GPU for computations can provide a performance boost. Cloud-based applications may not require use of any optimisations and may use it directly in real-time. Furthermore, since these results are generated from a personal computer, the technique may perform more efficiently in servers, which usually have better hardware specifications than a personal computer.

## VI. CONCLUSION

The technique classifies a pixel as skin only if the probability of neighbouring pixels being skin is sufficiently high. It also attempts to solve many limitations posed by existing techniques. Since the Deep Neural Network used in our technique is trained on HSV colour space, it is almost independent of illumination changes, which gives better results in variety of lighting conditions. The results show that the technique provides better segmentation results than the existing solutions. Noise generated by the Deep Neural Network model during classification of pixels is greatly reduced. The results generated thus contain denser skin regions reducing the sparser regions. The technique uses Deep Neural Network model which is computation-intensive but employing optimisations can make this technique fit for use in real-time applications. We can certainly use this technique to efficiently segment skin areas and produce better results.